# Title: Machine Intelligence for Outcome Predictions of Trauma Patients During Emergency Department Care

**Authors:** Joshua D. Cardosi[1], Herman Shen[1]*, Jonathan I. Groner[3,4,5], Megan Armstrong[3,4], Henry Xiang[2,3,4]

**Affiliations:**

[1]Department of Mechanical and Aerospace Engineering, The Ohio State University, 201 W. 19th Avenue, Columbus, OH, 43210.

[2]Department of Pediatrics, The Ohio State University, 370 West 9th Avenue, Columbus, OH 43210.

[3]Center for Pediatric Trauma Research; [4]Center for Pediatric Trauma Research, The Abigail Wexner Research Institute at Nationwide Children's Hospital, 700 Children's Drive, Columbus, OH 43205.

[5]Department of Surgery, The Ohio State University, 395 W 12th Ave #670, Columbus, OH 43210.

*To whom correspondence should be addressed: shen.1@osu.edu.

**One Sentence Summary:** A novel machine intelligence approach to predicting trauma patient mortality in the emergency department.

**Abstract**: Trauma mortality results from a multitude of non-linear dependent risk factors including patient demographics, injury characteristics, medical care provided, and characteristics of medical facilities; yet traditional approach attempted to capture these relationships using rigid regression models. We hypothesized that a transfer learning-based machine learning algorithm could deeply understand a trauma patient's condition and accurately identify individuals at high risk for mortality without relying on restrictive regression model criteria. Anonymous patient visit data were obtained from years 2007-2014 of the National Trauma Data Bank. Patients with incomplete vitals, unknown outcome, or missing demographics data were excluded. All patient visits occurred in U.S. hospitals, and of the 2,007,485 encounters that were retrospectively examined, 8,198 resulted in mortality (0.4%). The machine intelligence model was evaluated on its sensitivity, specificity, positive and negative predictive value, and Matthews Correlation Coefficient. Our model achieved similar performance in age-specific comparison models and generalized well when applied to all ages simultaneously. While testing for confounding factors, we discovered that excluding fall-related injuries boosted performance for adult trauma patients; however, it reduced performance for children. The machine intelligence model described here demonstrates similar performance to contemporary machine intelligence models without requiring restrictive regression model criteria or extensive medical expertise.

**Introduction**

Each year, thousands of trauma physicians and other health care personnel who provide emergency medical service care (EMSC) for injured patients face a critical clinical decision: to predict which injured patients will require immediate resources and interventions during the prehospital and initial emergency department (ED) care in order to prevent major complications or death. Out of the 139 million visits to the ED in 2017, approximately 278,000 resulted in patient mortality during transfer to or treatment in the ED (1). Common evidence-based tools such as Injury Severity Score (ISS) (2) can mislead medical professionals into the undertriage of patients or incorrectly classifying a patient's condition as unsurvivable (3). Additionally, regression model prediction techniques are often limited by restrictive mathematic model criteria. With the annual increase of ED visits outpacing the growth of the U.S. population (4), a more sophisticated prognostic tool will be necessary to achieve better patient outcomes by reducing undertriage and avoid unnecessary resource utilization resulting from the overtriage of trauma patients.

In the past 30 years, many researchers have sought to improve this clinical decision-making process and, more recently, automated prognostic tools that elucidate the seriousness of a patient's condition in a broad application setting. In 1992, McGonigal et al. demonstrated a groundbreaking neural network that relied only on Revised Trauma Score (RTS), ISS, and patient age and was capable of outperforming contemporary logistic regression models (5). Some years later, in 1999, Marble & Healy produced a more complicated model which could diagnose a patient with sepsis with almost 100% accuracy (6). However, while the results of these studies were impressive, they were only valid for a small subset of patients—in both cases they narrowed their focus to a specific type of injury. Since the publication of these papers, significant advancements in the field of machine intelligence have been made and a more ardent research effort than ever before is pushing towards modeling techniques that are valid across all patients, regardless of injury mechanism.

Several recent papers have demonstrated the power of neural networks in predicting patient health conditions and outcomes, but were formulated without an abundance of nationally representative data points, with model restrictions based on efforts specific to certain age groups, and without the verification of whether their models' performance was invariant across injury mechanisms (7-9). While these models showed great promise, the aforementioned issues created a gap in clinical understanding about their generalizability across patient demographics and conditions. There is, therefore, a need to study the capabilities of machine intelligence techniques on a sufficiently large and diverse data set with a direct focus on generalizability across clinical scenarios. Additionally, no study to date has utilized machine intelligence methods to predict death alone in the ED, despite the clinical relevance of such a risk assessment tool in prioritizing critical patients.

Using nationally representative of the trauma population, we aimed to develop a machine intelligence algorithm that could better predict patient mortality. Our hypothesis was that, with a sufficiently large data set that captures patient visit information from across the United States, an all-ages, injury-invariant, generalizable machine intelligence model can be built to better predict patient mortality in the ED than current practices. To validate the strength of the machine intelligence models across different age groups, we examined contemporary mortality prediction models, compared key performance metrics, and performance metrics for different injury inclusion criteria to ensure model invariance.

**Materials and Methods**

*Subhead 1: Study setting*

This was a retrospective study using 2007-2014 National Trauma Data Bank (NTDB) data. The American College of Surgeons (ACS) collects trauma registry data from hospitals across the U.S. every year and compiles it into the NTDB. Importantly, the ACS provides quality assurance through their National Trauma Data Standard (NTDS) Data Dictionary, which ensures the validity of data used by researchers (7).

*Subhead 2: Study samples*

From the 5.8 million patient visits captured in the data, we sought out trauma patient visits who had complete ED vitals, a known mode of arrival and transfer status, and who had a valid outcome (i.e. any disposition that was not 'not applicable', 'not known/recorded', or 'left against medical advice'). Any patients not meeting these criteria were removed from the data set.

*Subhead 3: Predictor variables*

We considered a total of 86 different predictors for mortality, all of which are typically available at the time of patient check-in and triage in the ED. Specifically, the following variables were chosen as predictors for our model: Abbreviated Injury Scale (AIS) (8), ISS (2), Glasgow Coma Scores (GCS) (9), NTDS comorbidities and external injury codes, whether the patient arrived by ambulance, whether the patient was transferred from another hospital, age, race, gender, and vitals collected at check-in (oxygen saturation, pulse, respiratory rate, temperature, and systolic blood pressure). As the NTDS has changed over the years, certain comorbidities were recorded in some years and not in others. Any chronic conditions that were not present across all years were removed from the data set. Additionally, the NTDS External Injury Codes were transformed into injury type, injury mechanism, and injury intent, based on the ICD-9-CM code recorded for the patient.

*Subhead 4: Outcome variables*

The outcome variable being predicted was patient mortality. Patients with a disposition of 'deceased/expired,' 'expired,' or 'discharged/transferred to hospice care' were treated as positive cases for patient mortality (10). All other valid outcomes were treated as negative for patient mortality. These included general admission to the hospital, admission to a specialized unit within the hospital (intensive care unit (ICU), step-down, etc.), transfer to another hospital, or discharge from the ED.

*Subhead 5: Model generation*

Pre-processing, a critical task in data mining, was required before the data could be passed to the machine intelligence model for training or prediction. Two separate, non-overlapping data sets were constructed; one contained hospital outcomes and the other contained ED outcomes. For each, a training set was created using 70% of the data available, and the remaining 30% was retained as a test set. As there are relatively few mortalities, these splits were created with a technique called stratification, which samples from the pool of mortality and non-mortality cases individually to ensure each class is represented proportionally. Additional data preparation included one hot encoding of categorical information and standardization of numerical data. Standardization was completed using Scikit-Learn's Standard Scaler, which was fitted to the training data and then used to transform both data splits (11).

The model architecture shown in Figure 1 was created using PyTorch and was composed of 4 distinct layers; it consisted of a single input layer, two hidden layers with 300 and 100 neurons, respectively, and a final output layer which output the model's prediction of patient mortality (12). Because the model contained thousands of connections between neurons, it could find complex non-linear relationships between different variables, but this also introduced the potential for overfitting on the training set. Several measures were taken to prevent model overfitting, such as batch normalization and dropout layers between neurons. This model architecture was applied to three different age groups: children only, adults only, and all ages.

Because of the scarcity of ED mortality data points, we combined transfer learning and synthetic data generation techniques to boost the discriminatory capabilities of the model (13, 14). First, the model was trained on the data set containing only hospital outcomes using a coarse learning rate. Then, using a finer learning rate and synthesized mortality data, the model was trained again on the data set containing ED outcomes.

The model's overall performance was gauged by calculating its sensitivity, specificity, sensitivity-specificity gap, area under receiver operating characteristic curve (AUC), positive predictive value (PPV), negative predictive value (NPV), and Matthews Correlation Coefficient (MCC). The sensitivity-specificity gap is the distance between these two values and explains how far the model is from having perfect predictive capabilities. It is calculated as shown in Equation 1.

$$\text{Gap} = (1 - \text{Sensitivity}) + (1 - \text{Specificity})$$

*Equation 1: Sensitivity-Specificity Gap*

MCC is a balanced measure between true positives, false positives, true negatives, and false negatives whereby the only means of improving the metric is reducing the total number of misclassifications. It is calculated as shown in Equation 2, where TP is true positives, TN is true negatives, FP is false positives, and FN is false negatives.

$$\text{MCC} = \frac{(TP * TN) - (FP * FN)}{\sqrt{(TP + FP)(TP + FN)(TN + FP)(TN + FN)}}$$

*Equation 2: Matthews Correlation Coefficient*

*Subhead 6: Model verification*

To validate our model's performance across all included patient visits, we collected results from other machine intelligence-based outcome prediction tools and compared our performance metrics. Goto et al. and Raita et al.'s models sought to predict either patient mortality or admission to the ICU with ED check-in data, while Hong et al. utilized triage data to predict whether a patient would be hospitalized (7-9). These techniques did not report a value for MCC and, for this reason, the metric has been left blank for those works. These works were selected due to their recency and utilization of modern machine intelligence methodologies. Because no contemporary machine intelligence model has tried to generalize to all ages before, we segmented the types of models into their appropriate age groupings. Additionally, to check the model's competence in predicting outcomes irrespective of the nature of a patient's injury, injury mechanisms determined by the reported external injury code were systematically filtered out of the data before training and testing the model.

In order to verify the strength of the machine intelligence model architecture itself, we created a second set of models which predicted the overall outcome of a patient, whether in the hospital or in the ED. This was an important step in verifying the model due to the scarcity of ED mortality data points and relative abundance of hospital deaths.

*Subhead 7: Statistical analysis of excluded patients*

Because of the reduction of the data set from 5.8 million patient visits to 2 million, we examined whether the patients which met our inclusion criteria occupied the same distribution as those who did not. We applied Student's t-test to the patient age, GCS Total, and ISS and the chi-square test to patient gender and presence of comorbidities. For each of these variables, we calculated a p-value with an alpha level of 0.05 to determine whether included and excluded patients were statistically similar. We discovered that the included patients are from a different distribution than those who were excluded, as all variables tested returned a p-value of zero.

**Results**

From 2007-2014, 5.8 million unique trauma patient visits were recorded in the NTDB with 1.7 million unique visits meeting our inclusion criteria. The data points which met these criteria was comprised of 300,847 children and 1,706,638 adults. Table 1 shows the characteristics of the child and adult populations with respect to the selected predictors and outcomes. From these selected data, the hospital outcome data set was constructed with 1,765,545 unique visits, and the ED outcome data set retained the remaining 245,940.

*Subhead 1: Model benchmarking`*

For children, our model achieved similar performance to Goto's Deep Neural Network (DNN) (15), with areas of improvement being PPV (0.08; 95% Confidence Interval (CI) 0.07-0.09) and NPV (0.993; 95% CI 0.991-0.995), as shown in Table 2. Across all other metrics, our model's performance characteristics fall within the confidence intervals given for the Goto DNN (15). Additionally, the size of our data set allows for our 95% CI to be much narrower than the comparison models for children.

When applied to adults alone, the model gives comparable performance to the comparison models. The sensitivity (0.75; 95% CI 0.75-0.75) is higher than the Hong Triage DNN (0.70) (16) and falls just below the Raita DNN (0.80; 95% CI 0.77-0.83) (17) while still achieving high specificity (0.84; 95% CI 0.84-0.84). While not the strongest in all aspects, the Sensitivity-Specificity Gap (0.41) still demonstrates that the model is the most balanced between sensitivity and specificity of the selected comparison models. No previous modeling effort has attempted to generalize across all age groups, so our all-ages model stands on its own in Table 2. Its performance metrics generally fall between those of our child and adult models, which suggests that the model has learned the relationship between a patient's age and whether certain health characteristics may result in mortality.

*Subhead 2: Performance across injury mechanisms*

The models for all-ages and adults-only both saw an increase in predictive performance across all metrics when fall injuries were removed from the test set. As shown in Table 3, the adult model without falls exhibited better AUC (0.82; 95% CI 0.81-0.83), specificity (0.91; 95% CI 0.91-0.91), sensitivity-specificity gap (0.35), PPV (0.20;95% CI 0.19-0.21), and MCC (0.673; 95% CI 0.666-0.680) while maintaining similar sensitivity (0.74; 95% CI 0.73-0.75) and NPV (0.991; 95% CI 0.990-0.992). The model for children appears to be weaker when falling injuries are excluded, with a lower sensitivity (0.76; 95% CI 0.75-0.77) and MCC (0.597; 95% CI 0.582-0.612). This implies that the predictor variables used in this study do not fully describe the condition of a patient who has suffered a fall. This process revealed that the model is invariant to all injury mechanisms in the NTDS except for falling injuries.

*Subhead 3: Architecture verification*

The second set of models, which predicted patients' overall outcome, outperformed the ED only models in most respects. For children, it achieved superior AUC (0.93; 95% CI 0.93-0.93), sensitivity (0.90 95% CI 0.90-0.90), specificity (0.97; 95% CI 0.97-0.97), sensitivity-specificity gap (0.13), PPV (0.19; 95% CI 0.19-0.19), NPV (0.999; 95% CI 0.999-0.999), and MCC (0.882; 95% CI 0.879-0.885), as can be seen in Table 4.

Similarly, for adults, the hospital & ED model achieved stronger AUC (0.85; 95% CI 0.85-0.85), sensitivity (0.90; 95% CI 0.90-0.90), sensitivity-specificity gap (0.30), NPV (0.997; 95% CI 0.997-0.997), and MCC (0.778; 95% CI 0.776-0.780). However, it did also feature weaker specificity (0.80; 95% CI 0.80-0.80) and PPV (0.11 95% CI 0.11-0.11).

Just like the ED only model, the hospital & ED all ages model achieved performance characteristics that indicated it had generalized for both children and adults. For all metrics but sensitivity, the model's performance fell between the models for children and adults.

**Discussion**

Implementation of our machine intelligence architecture on the NTDB provided innovative predictive capabilities that generalize to all trauma age groups and most types of injuries. With our data set of approximately 2 million unique visits, we created a single neural network architecture and trained three distinct models; we constructed unique models for children, adults, and all ages. Our models for children and adults achieved similar performance to the comparison models across most metrics, consistently ranking near the top for sensitivity, specificity, PPV, and NPV. The results suggest that the model architecture proposed here generalizes well across all ages. However, it is clear that fall injuries have the potential to confound the model, suggesting that the outcome of fall injuries may require more information than the included predictors provide.

It is important to note that another study not referenced in Table 2, the Trauma Quality Improvement Program (TQIP), has built a logistic regression model for child patient mortality that achieved an AUC of 0.996—almost perfect predictive power—but featured much narrower inclusion criteria than this study. Whereas the TQIP report limited their observations to victims of blunt, penetrating, or abuse-related injuries with at least one AIS of 2 or greater, we imposed none of these inclusion/exclusion criteria (18).

Our study is the largest of machine intelligence to date in trauma outcome prediction, featuring over 2 million unique patient encounters from across the United States. While previous studies demonstrated the promising capabilities of machine intelligence as a prognostic tool for patient outcomes, none captured the diverse healthcare settings that exist across the U.S. trauma patients, demonstrated invariance across injury mechanisms, or focused solely on patient mortality (7-9), and only one confirmed that additional data would not improve its model further (16). Machine intelligence is a famously data-driven technique, with few research groups having access to the amount of unique data points necessary to make their model the best it can be. With our large, diverse set of trauma data, we are confident that our model is not only the best it can be with its current architecture, but we also can say with high certainty that it will generalize to trauma patients all across the United States.

As we worked to ensure that our model's performance was invariant with respect to different injury mechanisms included in the NTDB, we discovered that the exclusion of fall injuries made a noticeable difference in the model's ability to discriminate between mortalities and survivors. It is well-known that adult fall injuries, especially in the elderly population, can result in hip fractures which lead to complications and death. Current triage guidelines acknowledge the complex nature of ground-level falls on the elderly, and at least one study has demonstrated that AIS and GCS are unreliable measures for assessing these patients' mortality risk levels (19, 20). The removal of these kinds of injuries then improved the performance of the adult model. However, the model for children achieved slightly worse performance without the inclusion of these fall injuries, indicating that the model can discern the seriousness of a child's fall-related injury well. Further investigation will be necessary to find the data and machine intelligence architecture that will help overcome this confounding factor.

The model architecture verification process in our study strongly indicates that the model architecture of Figure 1 can give predictions highly correlated with a patient's true outcome. The challenge in achieving reliable results for ED only cases lies in the scarcity of emergency department mortality data points, not with the modeling approach. Widening the inclusion criteria may allow for more training examples to be retained, but it will come at the cost of data richness.

The main limitation of this study is the need for a complete set of patient vitals. Our data set had approximately 5.8 million unique patient visits, but after filtering out data based on our inclusion criteria, we were left with 2 million. While this number of unique visits is sufficient for training the model, it signifies that there are many clinical scenarios that our machine intelligence architecture cannot handle. A trauma patient could be missing, for example, comorbidity information because their condition is too urgent to ask them about their chronic conditions. A distinct advantage of our study is the very broad inclusion criteria used. It is very common for medical research to limit patient inclusion to a very specific subset of all the combination of patient characteristics possible. While this specialization tends to give great performance, it also disallows the observations of the research from being applied to patients outside of the inclusion criteria. Our methods only filter out patients who are missing important information, such as vitals or demographics. We do not filter by age, injury mechanism, or any other categorical value, and we demonstrate that this is a viable approach in predicting patient outcomes.

Additionally, the NTDB provides a variety of medical facility-related information which may be pertinent in determining a patient's outcome but was not used in this study so that a fair assessment of our model could be made with respect to contemporary works, as they did not have access to facility variables. Some facilities, like level 1 trauma centers, will be better equipped than others to handle certain types of patients, and that reality is not captured in this study. Our rationale was to base our machine intelligence on patients demographics and injury characteristics so pre-hospital emergency medical service could use the prediction to guide trauma patient field triage . Finally, the anonymous nature of the data used means that our model can only analyze the outcomes of individual visits rather than the patients themselves. A longitudinal study would most likely benefit the model, as it could learn the patterns which contribute to patient deterioration over the long-term rather than during a single visit.

*Subhead 1: Future work*

Further research into the defining patient characteristics, model architecture, or pre-processing pipeline which allows the model to differentiate between fatal and survivable fall injuries is a necessary next step. This will address the performance loss observed when patients who have suffered a fall injury are included in the test set. Additionally, data related to the healthcare facility should be integrated into the predictive model, as this will help discern whether the patient can receive the care necessary to prevent mortality. Finally, the predictor variables selected for this study should be pruned to only include those which aid the model's performance, as this will result in fewer excluded patients and, therefore, more examples of patient mortality for the model to learn from.

*Subhead 2: Conclusion*

A predictive model for trauma patient mortality from approximately 2 million unique visits to the emergency department was developed and achieved best in class sensitivity-specificity gap, while simultaneously maintaining sensitivity and specificity similar to contemporary models. However, we also discovered that the predictors used in this study do not allow the model to fully differentiate between fatal and survivable fall injuries, as the model saw a significant performance boost when fall injuries were removed from the data set. Future work will need to determine the predictors or processing methods needed to overcome this confounding factor. Ultimately, though, this study demonstrates that machine intelligence models are capable of giving predictions highly correlated with a trauma patient's true outcome, and as a result, healthcare workers in the emergency department may use them as a risk assessment aid when determining the urgency of a patient's condition. This will reduce the burden on healthcare personnel, prevent over-utilization of resources due to overtriage, and improve the quality of care available to those who truly need it to reduce mortality risk.

**References and Notes:**

1. National Center for Health Statistics, "National Hospital Ambulatory Medical Care Survey: 2017 Emergency Department Summary Tables," Centers for Disease Control and Prevention, 2017.

2. S. P. Baker, B. O'Neill, W. Haddon, W. B. Long, The Injury Severity Score: a method for describing patients with multiple injuries and evaluating emergency care. *The Journal of Trauma,* vol. 14, no. 3, pp. 187-196 (1974).

3. L. B. Elgin, S. J. Appel, D. Grisham, S. Dunlap, Comparisons of Trauma Outcomes and Injury Severity Score. *Journal of Trauma Nursing,* vol. 26, no. 4, pp. 199-207 (2019).

4. M. J. Brian, C. Stocks, P. L. Owens, Trends in Emergency Department Visits, 2006-2014. HCUP (2017).

5. M. D. McGonigal, J. Cole, W. Schwab, D. R. Kauder, M. F. Rotondo, P. B. Angood, A New Approach to Probability of Surviving Score for Trauma Quality Assurance. *The Journal of Trauma,* pp. 863-870 (1992).

6. R. P. Marble, J. C. Healy, A neural network approach to the diagnosis of morbidity outcomes in trauma care. *Artificial Intelligence in Medicine,* pp. 299-307 (1999).

7. "About NTDB," 2020. [Online]. Available: https://www.facs.org/quality-programs/trauma/tqp/center-programs/ntdb/about.

8. Association for the Advancement of Automotive Medicine, "Abbreviated Injury Scale (AIS)," Association for the Advancement of Automotive Medicine, 2020. [Online]. Available: https://www.aaam.org/abbreviated-injury-scale-ais/. [Accessed 19 August 2020].

9. B. Jennett, G. Teasdale, Assessment of coma and impaired consciousness. A practical scale. *The Lancet,* vol. 304, no. 7872, pp. 81-84 (1974).

10. A. H. Kaji, Z. G. Hashmi, A. B. Nathens, Practical Guide to Surgical Data Sets: National Trauma Data Bank (NTDB). *JAMA Surgery* (2018).

11. F. Pedregosa, G. Varoquaux, A. Gramfort, V. Michel, B. Thirion, O. Grisel, M. Blondel, P. Prettenhofer, R. Weiss, V. Dubourg, J. Vanderplas, A. Passos, D. Cournapeau, M. Brucher, M. Perrot, E. Duchesnay, "Scikit-learn: Machine Learning in Python," *JMLR 12,* pp. 2825-2830 (2011).

12. A. Paszke, S. Gross, F. Massa, A. Lerer, J. Bradbury, G. Chanan, T. Killeen, Z. Lin, N. Gimelshein, L. Antiga, A. Desmaison, A. Kopf, E. Yang, Z. DeVito, M. Raison, A. Tejani, S. Chilamkurthy, B. Steiner, L. Fang, J. Bai, S. Chintala, PyTorch: An Imperative Style, High-Performance Deep Learning Library. *NIPS*, Vancouver (2019).

13. L. Y. Pratt, *NIPS Conference: Advances in Neural Information Processing Systems 5* (1993), pp. 204-211.

14. G. Lemaitre, F. Nogueira, C. K. Aridas, Imbalanced-learn: A Python Toolbox to Tackle the Curse of Imbalanced Datasets in Machine Learning. *Journal of Machine Learning Research,* vol. 18, no. 17, pp. 1-5 (2017).

15. T. Goto, C. A. Camargo, M. K. Faridi, R. J. Freishtat, K. Hasegawa, Machine Learning–Based Prediction of Clinical Outcomes for Children During Emergency Department Triage. *JAMA Network Open* (2019).

16. W. S. Hong, A. D. Halmovich, A. Taylor, Predicting hospital admission at emergency department triage using machine learning. *PLOS ONE* (2018).

17. Y. Raita, T. Goto, M. K. Faridi, D. F. Brown, C. A. Camargo, K. Hasegawa, Emergency department triage prediction ofclinical outcomes using machine learning models. *BMC* (2019).

18. American College of Surgeons Committee on Trauma, ACS Pediatric TQIP Aggregate Report: Spring 2016. American College of Surgeons (2016).

19. Centers for Disease Control and Prevention, Guidelines for Field Triage of Injured Patients: Recommendations of the National Expert Panel on Field Triage, 2011. (2012).

20. S. R. Konda, A. Lott and K. A. Egol, The Coming Hip and Femur Fracture Bundle: A New Inpatient Risk Stratification Tool for Care Providers. *Geriatric Orthopaedic Surgery & Rehabilitation,* vol. 9 (2018).

**Funding**: No external funding was received for this study.
**Author contributions**:
- JDC – Data analysis, machine intelligence model design, data interpretation, literature search, writing
- HS – Study design, machine intelligence model design, critical revision, literature search
- JIG – Critical revision, literature search
- MA – Critical revision, literature search
- HX – Study design, critical revision, literature search

**Competing interests**: All authors have no conflicts of interest or funding sources to disclose.
**Human Subjects Research Statement:** As this was a retrospective study, informed consent was not required.

**Figures**:

**Fig. 1**. Model Architecture and Sample Training Loss. The model consisted of 3 layers and utilized both batch normalization and dropout to smooth training loss and prevent overfitting of the model to the training set.

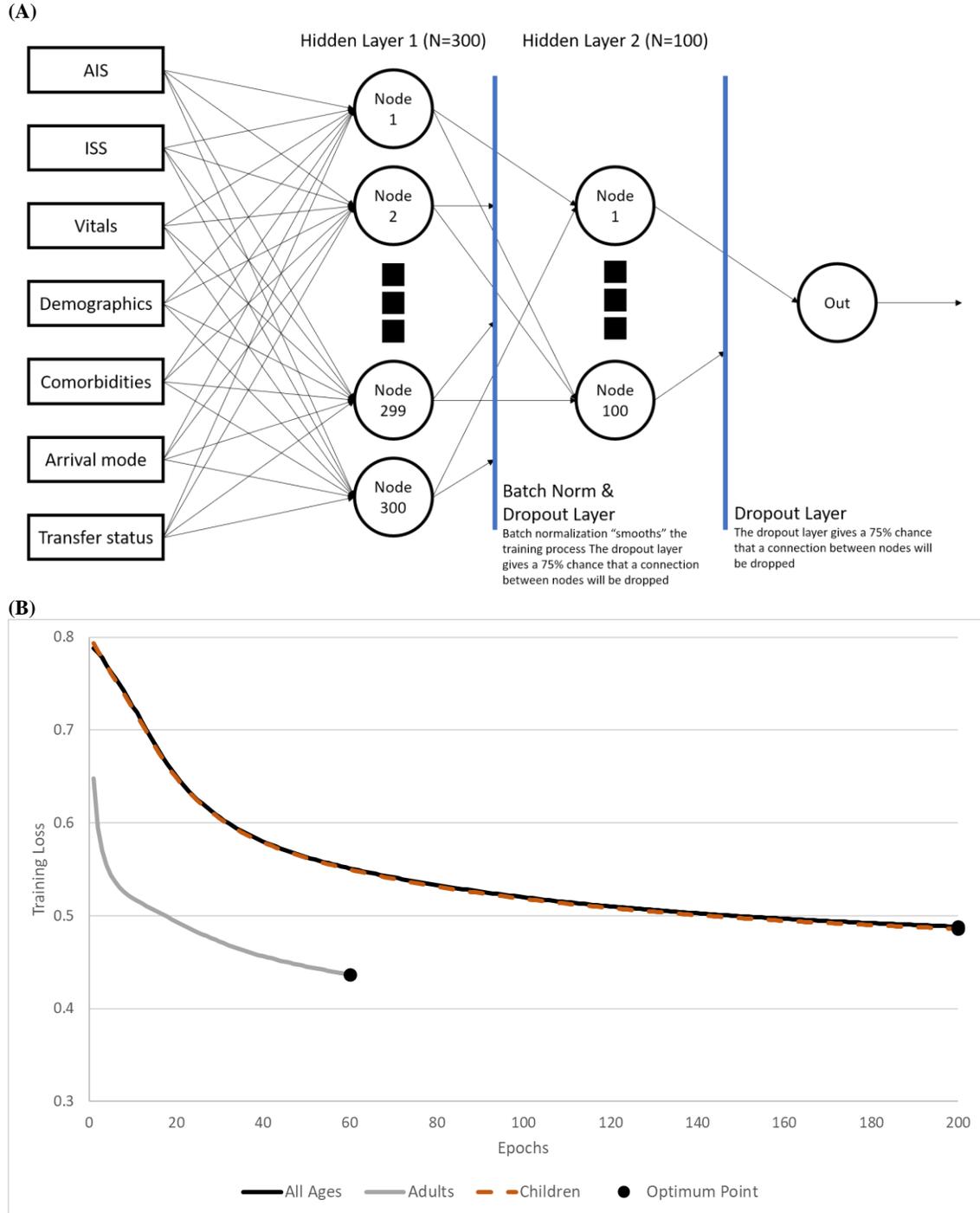

**Table 1.** Predictor and Trauma Outcome Variables.

| Variable | n = 300,847 children | | n = 1,706,638 adults | |
|---|---|---|---|---|
| **Demographics** | | | | |
| Age (year), mean (Std. Dev) | 10.42 | (5.91) | 51.67 | (20.93) |
| Female Sex | 99,523 | (33.92) | 632,346 | (37.95) |
| White | 196,736 | (67.06) | 1,258,913 | (75.54) |
| Black or African American | 51,994 | (17.72) | 229,302 | (13.76) |
| Other Race | 34,196 | (11.66) | 123,493 | (7.41) |
| Asian | 5,079 | (1.73) | 27,888 | (1.67) |
| American Indian | 3,348 | (1.14) | 14,902 | (0.89) |
| Race N/A | 1,170 | (0.4) | 8,551 | (0.51) |
| Native Hawaiian or Other Pacific Islander | 853 | (0.29) | 3,411 | (0.2) |
| **ED Vitals** | | | | |
| Oxygen Saturation, mean (Std. Dev.) | 98.26 | (6.93) | 96.85 | (7.44) |
| Systolic Blood Pressure, mean (Std. Dev.) | 122.48 | (19.53) | 139.89 | (26.35) |
| Pulse, mean (Std. Dev.) | 102.42 | (26.18) | 87.49 | (19.13) |
| Respiratory Rate, mean (Std. Dev.) | 21.32 | (6.82) | 18.40 | (4.63) |
| Temperature, mean (Std. Dev.) | 36.67 | (1.26) | 36.52 | (1.45) |
| GCS Eye, mean (Std. Dev.) | 3.85 | (0.62) | 3.84 | (0.64) |
| GCS Verbal, mean (Std. Dev.) | 4.75 | (0.87) | 4.69 | (0.91) |
| GCS Motor, mean (Std. Dev.) | 5.79 | (0.91) | 5.77 | (0.96) |
| Injury Severity Score, mean (Std. Dev.) | 7.36 | (7.21) | 9.08 | (7.82) |
| AIS Area 1, mean (Std. Dev.) | 1.11 | (1.73) | 1.09 | (1.77) |
| AIS Area 2, mean (Std. Dev.) | 0.31 | (0.63) | 0.34 | (0.67) |
| AIS Area 3, mean (Std. Dev.) | 0.03 | (0.36) | 0.04 | (0.33) |
| AIS Area 4, mean (Std. Dev.) | 0.34 | (1.00) | 0.62 | (1.26) |
| AIS Area 5, mean (Std. Dev.) | 0.30 | (0.95) | 0.24 | (0.82) |
| AIS Area 6, mean (Std. Dev.) | 0.19 | (0.69) | 0.43 | (0.96) |
| AIS Area 7, mean (Std. Dev.) | 0.60 | (0.96) | 0.52 | (0.89) |
| AIS Area 8, mean (Std. Dev.) | 0.60 | (1.07) | 0.87 | (1.21) |
| AIS Area 9, mean (Std. Dev.) | 0.13 | (0.42) | 0.10 | (0.37) |
| **Comorbidities** | | | | |
| Alcoholism | 2,420 | (0.82) | 152,602 | (9.16) |
| Angina | 8 | (0) | 3,873 | (0.23) |
| Ascites within 30 days | 55 | (0.02) | 1,265 | (0.08) |
| Bleeding Disorder | 668 | (0.23) | 96,168 | (5.77) |
| Chemotherapy | 58 | (0.02) | 4,358 | (0.26) |
| Congenital Anomalies | 2,248 | (0.77) | 4,598 | (0.28) |
| Congestive heart failure | 92 | (0.03) | 57,005 | (3.42) |
| Current smoker | 10,098 | (3.44) | 315,492 | (18.93) |
| CVA/residual neurological deficit | 233 | (0.08) | 38,609 | (2.32) |
| Diabetes mellitus | 1,074 | (0.37) | 209,902 | (12.6) |
| Disseminated cancer | 34 | (0.01) | 11,608 | (0.7) |
| Esophageal varices | 48 | (0.02) | 3,924 | (0.24) |
| Functionally dependent health status | 651 | (0.22) | 32,606 | (1.96) |
| Hypertension requiring medication | 1,054 | (0.36) | 527,251 | (31.64) |
| Myocardial Infarction | 19 | (0.01) | 23,487 | (1.41) |
| No comorbidities | 199,555 | (68.02) | 442,939 | (26.58) |
| Obesity | 3,684 | (1.26) | 110,593 | (6.64) |
| Prematurity | 1,666 | (0.57) | 412 | (0.02) |
| PVD | 14 | (0) | 7,831 | (0.47) |
| Respiratory Disease | 16,312 | (5.56) | 137,284 | (8.24) |
| Steroid use | 109 | (0.04) | 8,746 | (0.52) |
| **Injury Intent** | | | | |
| Assault | 21,319 | (7.27) | 177,543 | (10.65) |
| Other | 228 | (0.08) | 3,098 | (0.19) |
| Self-inflicted | 2,449 | (0.83) | 25,798 | (1.55) |
| Undetermined | 1,820 | (0.62) | 6,135 | (0.37) |
| Unintentional | 265,434 | (90.48) | 1,447,143 | (86.84) |
| **Injury Type** | | | | |
| Blunt | 241,153 | (82.2) | 1,427,232 | (85.64) |
| Burn | 9,740 | (3.32) | 25,843 | (1.55) |
| Other/unspecified | 21,498 | (7.33) | 65,390 | (3.92) |
| Penetrating | 18,859 | (6.43) | 141,252 | (8.48) |
| **Injury Mechanism** | | | | |
| Adverse effects, drugs | 34 | (0.01) | 307 | (0.02) |
| Adverse effects, medical care | 22 | (0.01) | 406 | (0.02) |
| Cut/pierce | 8,855 | (3.02) | 77,958 | (4.68) |
| Drowning/submersion | 269 | (0.09) | 650 | (0.04) |
| Fall | 95,199 | (32.45) | 690,746 | (41.45) |
| Fire/flame | 2,874 | (0.98) | 15,864 | (0.95) |
| Firearm | 9,982 | (3.4) | 63,173 | (3.79) |
| Hot object/substance | 6,866 | (2.34) | 9,979 | (0.6) |
| MVT Motorcyclist | 3,968 | (1.35) | 91,688 | (5.5) |
| MVT Occupant | 51,335 | (17.5) | 334,239 | (20.06) |
| MVT Other | 932 | (0.32) | 3,423 | (0.21) |
| MVT Pedal cyclist | 4,436 | (1.51) | 13,306 | (0.8) |
| MVT Pedestrian | 12,745 | (4.34) | 47,643 | (2.86) |
| MVT Unspecified | 457 | (0.16) | 4,193 | (0.25) |
| Machinery | 1,101 | (0.38) | 20,115 | (1.21) |
| Natural/environmental, Bites and stings | 4,949 | (1.69) | 7,497 | (0.45) |
| Natural/environmental, Other | 1,543 | (0.53) | 5,371 | (0.32) |
| Other specified and classifiable | 8,773 | (2.99) | 21,390 | (1.28) |
| Other specified, not elsewhere classifiable | 1,379 | (0.47) | 7,560 | (0.45) |
| Overexertion | 1,490 | (0.51) | 4,715 | (0.28) |
| Pedal cyclist, other | 11,880 | (4.05) | 25,423 | (1.53) |
| Pedestrian, other | 1,568 | (0.53) | 4,819 | (0.29) |
| Poisoning | 330 | (0.11) | 647 | (0.04) |
| Struck by, against | 32,068 | (10.93) | 111,135 | (6.67) |
| Suffocation | 281 | (0.1) | 1,396 | (0.08) |
| Transport, other | 25,462 | (8.68) | 80,501 | (4.83) |
| Unspecified | 2,452 | (0.84) | 15,573 | (0.93) |
| **Arrived by Ambulance** | 223,432 | (76.16) | 1,409,459 | (84.58) |
| **Transferred from Other Hospital** | 108,720 | (37.06) | 387,609 | (23.26) |
| **Mortality** | 1,053 | (0.36) | 7,145 | (0.43) |

Unless otherwise noted, data are presented as count (percentage) of positive cases

**Table 2.** Predictor and Trauma Outcome Variables.

| Model | AUC (95% CI) | Sensitivity (95% CI) | Specificity (95% CI) | Gap* | PPV (95% CI) | NPV (95% CI) | MCC (95% CI) |
|---|---|---|---|---|---|---|---|
| *Models for Children* | | | | | | | |
| Goto LR (15) | 0.78 (0.71-0.85) | 0.54 (0.39-0.69) | 0.91 (0.75-0.93) | 0.55 | 0.01 (0.01-0.02) | 0.990 (0.990-0.990) | - |
| Goto DNN (15) | 0.85 (0.78-0.92) | 0.78 (0.63-0.90) | 0.77 (0.62-0.92) | 0.45 | 0.01 (0.01-0.02) | 0.990 (0.990-0.990) | - |
| Ours | 0.78 (0.77-0.79) | 0.80 (0.79-0.81) | 0.75 (0.74-0.76) | 0.45 | 0.08 (0.07-0.09) | 0.993 (0.991-0.995) | 0.633 (0.621-0.645) |
| *Models for Adults* | | | | | | | |
| Raita LR (17) | 0.74 (0.72-0.75) | 0.50 (0.47-0.53) | 0.86 (0.82-0.87) | 0.64 | 0.07 (0.05-0.08) | 0.988 (0.988-0.988) | - |
| Raita DNN (17) | 0.86 (0.85-0.87) | 0.80 (0.77-0.83) | 0.76 (0.73-0.78) | 0.44 | 0.06 (0.06-0.07) | 0.995 (0.994-0.995) | - |
| Hong Triage DNN (16) | 0.87 (0.87-0.88) | 0.70 | 0.85 | 0.45 | 0.66 | 0.870 | - |
| Ours | 0.80 (0.80-0.80) | 0.75 (0.75-0.75) | 0.84 (0.84-0.84) | 0.41 | 0.13 (0.13-0.13) | 0.991 (0.990-0.992) | 0.645 (0.640-0.650) |
| *Model for All Ages* | | | | | | | |
| Ours | 0.78 (0.78-0.78) | 0.72 (0.72-0.72) | 0.84 (0.84-0.84) | 0.44 | 0.12 (0.12-0.12) | 0.990 (0.989-0.991) | 0.601 (0.596-0.606) |

**Table 3.** Comparison of Performance with and without Fall Injuries.

| Model | AUC (95% CI) | Sensitivity (95% CI) | Specificity (95% CI) | Gap* | PPV (95% CI) | NPV (95% CI) | MCC (95% CI) |
|---|---|---|---|---|---|---|---|
| *Model for Children* | | | | | | | |
| With falls | 0.78 (0.77-0.79) | 0.80 (0.79-0.81) | 0.75 (0.74-0.76) | 0.45 | 0.08 (0.07-0.09) | 0.993 (0.991-0.995) | 0.633 (0.621-0.645) |
| No falls | 0.76 (0.75-0.77) | 0.76 (0.75-0.77) | 0.76 (0.75-0.77) | 0.48 | 0.06 (0.05-0.07) | 0.993 (0.990-0.996) | 0.597 (0.582-0.612) |
| *Model for Adults* | | | | | | | |
| With falls | 0.80 (0.80-0.80) | 0.75 (0.75-0.75) | 0.84 (0.84-0.84) | 0.41 | 0.13 (0.13-0.13) | 0.991 (0.990-0.992) | 0.645 (0.640-0.650) |
| No falls | 0.82 (0.81-0.83) | 0.74 (0.73-0.75) | 0.91 (0.91-0.91) | 0.35 | 0.20 (0.19-0.21) | 0.991 (0.990-0.992) | 0.673 (0.666-0.680) |
| *Model for All Ages* | | | | | | | |
| With falls | 0.78 (0.78-0.78) | 0.72 (0.72-0.72) | 0.84 (0.84-0.84) | 0.44 | 0.12 (0.12-0.12) | 0.990 (0.989-0.991) | 0.601 (0.596-0.606) |
| No falls | 0.83 (0.83-0.83) | 0.76 (0.75-0.77) | 0.90 (0.90-0.90) | 0.34 | 0.18 (0.18-0.18) | 0.992 (0.991-0.993) | 0.688 (0.682-0.694) |

*The gap between sensitivity and specificity. Calculated as follows: Gap = (1-Sensitivity) + (1-Specificity).

**Table 4.** Model Performance for Varying Outcome Predictions.

| Model | AUC (95% CI) | Sensitivity (95% CI) | Specificity (95% CI) | Gap* | PPV (95% CI) | NPV (95% CI) | MCC (95% CI) |
|---|---|---|---|---|---|---|---|
| *Model for Children* | | | | | | | |
| ED Only | 0.78 (0.77-0.79) | 0.80 (0.79-0.81) | 0.75 (0.74-0.76) | 0.45 | 0.08 (0.07-0.09) | 0.993 (0.991-0.995) | 0.633 (0.621-0.645) |
| Hospital & ED | 0.93 (0.93-0.93) | 0.90 (0.90-0.90) | 0.97 (0.97-0.97) | 0.13 | 0.19 (0.19-0.19) | 0.999 (0.999-0.999) | 0.882 (0.879-0.885) |
| *Model for Adults* | | | | | | | |
| ED Only | 0.80 (0.80-0.80) | 0.75 (0.75-0.75) | 0.84 (0.84-0.84) | 0.41 | 0.13 (0.13-0.13) | 0.991 (0.990-0.992) | 0.645 (0.640-0.650) |
| Hospital & ED | 0.85 (0.85-0.85) | 0.90 (0.90-0.90) | 0.80 (0.80-0.80) | 0.30 | 0.11 (0.11-0.11) | 0.997 (0.997-0.997) | 0.778 (0.776-0.780) |
| *Model for All Ages* | | | | | | | |
| ED Only | 0.78 (0.78-0.78) | 0.72 (0.72-0.72) | 0.84 (0.84-0.84) | 0.44 | 0.12 (0.12-0.12) | 0.990 (0.989-0.991) | 0.601 (0.596-0.606) |
| Hospital & ED | 0.86 (0.86-0.86) | 0.91 (0.91-0.91) | 0.81 (0.81-0.81) | 0.28 | 0.11 (0.11-0.11) | 0.997 (0.997-0.997) | 0.797 (0.795-0.799) |

*The gap between sensitivity and specificity. Calculated as follows: Gap = (1-Sensitivity) + (1-Specificity).